\documentclass{article}

\usepackage{microtype}
\usepackage{graphicx}
\usepackage{subcaption}
\usepackage{booktabs}
\usepackage{multirow}
\usepackage{hyperref}

\usepackage[preprint]{icml2026}

\usepackage{amsmath}
\usepackage{amssymb}
\usepackage{mathtools}
\usepackage{amsthm}
\usepackage{tabularx}

\usepackage[capitalize,noabbrev]{cleveref}
\theoremstyle{plain}

\theoremstyle{definition}

\theoremstyle{remark}

\usepackage[textsize=tiny]{todonotes}

\newcommand{\model}{3DSPA }

\icmltitlerunning{3D Semantic Point Autoencoder (3DSPA)}

\begin{document}

\twocolumn[
  \icmltitle{3DSPA: A 3D Semantic Point Autoencoder for Evaluating Video Realism}

  \icmlsetsymbol{equal}{*}

\begin{icmlauthorlist}
  \icmlauthor{Bhavik Chandna}{ucsd,vector}
  \icmlauthor{Kelsey R. Allen}{vector,ubc}
\end{icmlauthorlist}

\icmlaffiliation{ucsd}{University of California, San Diego, CA, USA}
\icmlaffiliation{ubc}{University of British Columbia, Vancouver, BC, Canada}
\icmlaffiliation{vector}{Vector Institute, Toronto, ON, Canada}

\icmlcorrespondingauthor{Bhavik Chandna}{bchandna@ucsd.edu}
\icmlcorrespondingauthor{Kelsey R. Allen}{kelsey.allen@vectorinstitute.ai}

  \icmlkeywords{Machine Learning, ICML}

  \vskip 0.3in
]

\printAffiliationsAndNotice{}

\begin{abstract}
AI video generation is evolving rapidly. For video generators to be useful for applications ranging from robotics to film-making, they must consistently produce realistic videos. However, evaluating the realism of generated videos remains a largely manual process -- requiring human annotation or bespoke evaluation datasets which have restricted scope.
Here we develop an automated evaluation framework for video realism  which captures both semantics and coherent 3D structure and which does not require access to a reference video. Our method, 3DSPA, is a 3D spatiotemporal point autoencoder which integrates 3D point trajectories, depth cues, and DINO semantic features into a unified representation for video evaluation. 3DSPA models how objects move and what is happening in the scene, enabling robust assessments of realism, temporal consistency, and physical plausibility. Experiments show that 3DSPA reliably identifies videos which violate physical laws, is more sensitive to motion artifacts, and aligns more closely with human judgments of video quality and realism across multiple datasets. Our results demonstrate that enriching trajectory-based representations with 3D semantics offers a stronger foundation for benchmarking generative video models, and implicitly captures physical rule violations. The code and pretrained model weights will be available at 
\href{https://github.com/TheProParadox/3dspa_code}{https://github.com/TheProParadox/3dspa\_code}.
\end{abstract}

\section{Introduction}

Recent years have witnessed rapid progress in generative video models, with systems such as Sora \cite{videoworldsimulators2024}, Veo \cite{veogoogle}, Kling AI \cite{kling2024}, and Luma-Ray \cite{luma2025ray2} capable of producing high-resolution, long-duration videos. These systems have started to showcase unprecedented visual fidelity, with coherent multi-object movement, smooth camera motion, across diverse scenes. However, the ultimate objective of developing these text-to-video models has always been to generate videos which are not only visually compelling but also realistic; capturing semantic meaning, temporal consistency, and physical plausibility in a way that mirrors a real-world video. If achieved, it will generate tremendous excitement across domains ranging from robotics and embodied AI \cite{wu2023unleashing, yang2025orv, fu2025learning} to virtual reality \cite{christian2025ai}, education \cite{xu2025recorded}, and creative industries like advertising and film-making.

\begin{figure}[!t]
    \centering
        \includegraphics[width=1\linewidth]{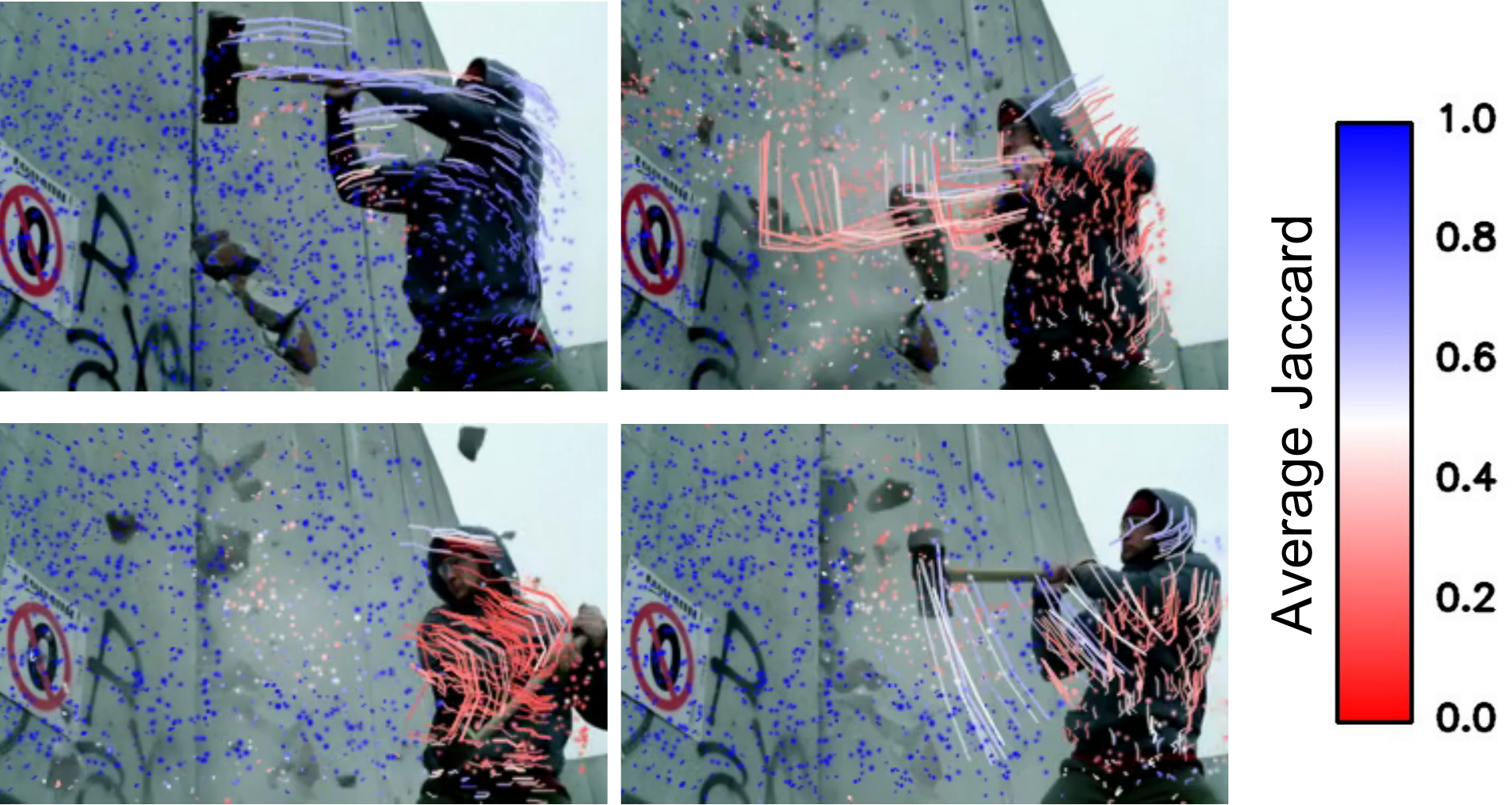}
        \caption{Example 3D point tracks (projected into 2D) reconstructed by \model for an unrealistic generated video from VideoPhy-2 (video frames progress from left to right, top to bottom) \cite{bansal2025videophy}, depicting a man striking a wall with a hammer and clearly violating physical laws. \model assigns low realism as measured with the Average Jaccard ($AJ$) (blue: high, red: low, white: intermediate). The video ranks among the lowest in both \model’s scores and human realism ratings, highlighting strong alignment with human judgment.}

        \label{fig:3dreconstruct}
\end{figure}
        
Understanding the realism of generated videos is more than an aesthetic problem. It directly affects their utility for various downstream applications. For example, in robotics and embodied AI, policies trained based on simulated video rollouts that are not physically plausible may not transfer successfully to real world deployments. Similarly, in entertainment, audiences are sensitive to subtle cues of unrealistic motion, which can undermine immersion. Thus, a systematic way of measuring whether generated videos are physically plausible and perceptually realistic is a foundational requirement for practical use across multiple settings.

However, existing approaches to measuring realism remain limited. The most common strategy is to rely on human annotation, where raters provide subjective assessments of qualities such as naturalness, temporal smoothness, and physical plausibility \cite{liu2024evalcrafter, bansal2025videophy}. While such annotations are informative, they are expensive and time-consuming, and do not scale to the vast number of videos modern generative systems can produce. A second line of work has attempted to build benchmarks by constructing datasets of paired real and fake videos \cite{borji2022pros, chen2024demamba, motamed2025generative, bear2023physionpp}. Yet this requires careful curation of datasets, often domain-specific, and assumes that generated samples are comparable to available real-world footage. Neither approach provides a scalable, general-purpose solution.

Moreover, prior automated measures have largely equated realism with temporal consistency, simply ensuring that videos do not exhibit frame-to-frame flickering or incoherence. While temporal smoothness is indeed important, it is not sufficient. Realism also requires adherence to the semantics of motion and the physical laws that govern objects in three dimensions. For example, a video where a ball bounces upward indefinitely without slowing down might look temporally smooth but is physically implausible. Likewise, a car turning a corner but sliding sideways without frictional constraints violates semantic expectations of how vehicles move. Prior work has struggled to capture such failures because they demand reasoning about both semantics and 3D structure, not just pixels over time. Existing automated evaluation metrics, e.g. \cite{allen2025direct},  operate in 2D feature spaces, neglecting the fact that real-world objects persist in three dimensions, maintain continuity across occlusion, and obey physical laws such as gravity, inertia, and collision.

To address these challenges, we propose \textbf{\model} (3D Semantic Point Autoencoder), a novel framework for assessing the realism of generated videos. \model combines semantic features with 3D point track auto-encoding. The key idea is to represent a video as a sequence of tracked 3D points, enriched with semantic embeddings, and train an autoencoder that reconstructs these tracks. By compressing and reconstructing motion trajectories, the model is forced to capture underlying physical and semantic regularities, making deviations from realism detectable.

The main contributions of our paper include:
\begin{itemize}
    \item First, we demonstrate that \model functions as a capable 3D point tracker, despite the information bottleneck inherent in auto-encoding. 
    \item Second, we show that it reliably detects violations of physical laws in controlled synthetic settings by using the IntPhys2 benchmark \cite{bordes2025intphys}.
    \item  Finally, we evaluate \model on two existing generated video datasets which provide human annotations of realism: EvalCrafter \cite{liu2024evalcrafter} and VideoPhy-2 \cite{bansal2025videophy}, and find that it better aligns with human judgments of motion quality and physical realism compared to existing baselines.
\end{itemize}
  Together, these results suggest that incorporating semantics and 3D structure is essential for scalable, automated evaluation of generative video realism.

\section{Related work}
\textbf{Physical benchmarking for video models} Benchmarking intuitive and broader physical understanding has been central to recent progress in machine perception. In simulation, earlier works include IntPhys \cite{riochet2018intphys} which evaluates models on core physical expectations through possible vs. impossible event videos and CLEVRER \cite{yi2020clevrer} which aimed to advance causal reasoning in videos via descriptive, explanatory, and counterfactual questions. Recent works include Physion++ \cite{bear2023physionpp} which creates visual physics prediction tasks involving rigid, soft, and fluid dynamics, and IntPhys2 \cite{bordes2025intphys} which introduces more complex backgrounds and objects to the original IntPhys. In the real world, InfLevel \cite{weihs2022benchmarking} extends these ideas by collecting real-world videos of physics violations. Physics IQ \cite{motamed2025generative} and TRAVL \cite{motamed2025travl} similarly provide a set of real-world videos covering a variety of physical principles. In these benchmarks, models' physical commonsense is measured by either directly comparing their predictions about how a real world video will unfold to the video itself, or by investigating whether models are more \emph{surprised} by an unreal video than by a real one.

\textbf{Video quality assessment}
Video generative models are rapidly progressing, but it is still unclear how far they are from being able to generate videos which are indistinguishable from reality. Even evaluating this progress remains tricky. Earlier methods such as FVD~\cite{unterthiner2018towards} and CLIP-based scores~\cite{radford2021learning} measure frame quality and text-frame alignment respectively but fail to capture realism more broadly, which jointly depends on semantic coherence, motion consistency, and physical plausibility. Recent efforts aim to create automated evaluators which tackle realism more directly, including physics-aware metrics~\cite{chen2025physical}, and large-scale benchmarks such as VBench \cite{huang2024vbench}. However, many generative models have started saturating these benchmarks, often achieving high scores of 90\%+, indicating that the metrics are simply not challenging enough. As a result, benchmarks such as EvalCrafter \cite{liu2024evalcrafter} and VideoPhy-2 \cite{bansal2025videophy} rely on human judgments for comprehensive evaluation, and therefore avoid issues of benchmark saturation. Automated evaluators in these settings include optical flow and fine-tuned vision-language models. Several recent works also explore detecting or characterizing synthetic videos through temporal artifacts. \citet{zheng2025d3} propose a training-free detector that exploits second-order temporal features derived from inter-frame differences. Similarly, \citet{shi2025tcmabnet} introduce a spatial-temporal network combining CNN-based spatial modeling with temporal dynamics to capture fine-grained forgery cues. While effective in synthetic video detection, these methods primarily target binary classification rather than aligning with human judgments of motion realism. \citet{allen2025direct} introduces a 2D point-trajectory autoencoder for automatic realism evaluation focusing on providing a new automated metric aligning with human judgments, but their approach primarily captures temporal consistency.

\begin{figure*}[!t]
    \centering
    \includegraphics[width=0.95\textwidth]{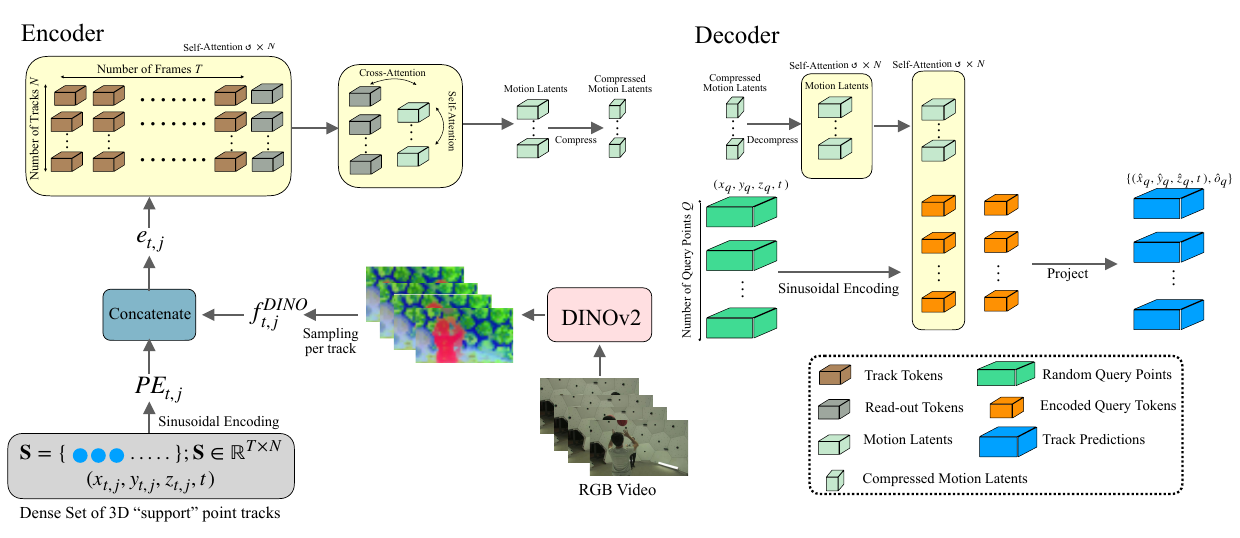}
    \caption{\textbf{\model architecture overview.} The encoder integrates 3D trajectories, temporal embeddings, and DINOv2 \cite{oquab2023dinov2} semantic features into a compact latent representation using occlusion-aware attention and a Perceiver-style transformer architecture \cite{jaegle2021perceiver}. The decoder conditions on query points to reconstruct full 3D trajectories with occlusion flags.} \label{fig:3dspa}
\end{figure*}

\section{Model}
Our goal is to provide an automated metric that can capture human ratings of realism for any video.
To that end, we introduce \model, a 3D Semantic Point Autoencoder.
\model can be viewed as an extension of TRAJAN \cite{allen2025direct}, a 2D point trajectory autoencoder that is trained to map a support set of point trajectories into a fixed-size motion latent representation which is further used to reconstruct query point tracks. While TRAJAN does a good job in capturing motion information in its latent space, it has some limitations. First, the model has no knowledge of the visual environment, only the point trajectories themselves. This means it cannot reason about scene context, object interactions, or occlusions that can be crucial for judging motion plausibility. Second, restricting motion to 2D trajectories is not sufficient for evaluating realistic dynamics, which naturally occur in 3D. As a result, TRAJAN cannot fully capture the complexity of real-world motion.
\model is instead designed to reconstruct 3D point tracks from random queries across space and time and provide a semantics-aware motion latent representation.

\subsection{Architecture}
\model adopts an encoder–decoder setup, where the encoder $E$ operates on a dense set of support point tracks $\mathbf{S} = \{s_{t,j}\}$, with each track defined as $s_{t,j} = (x_{t,j}, y_{t,j}, z_{t,j}, o_{t,j})$. Here, $(x, y, z)$ denote the 3D position and $o$ is a binary occlusion flag at time $t$ for the $j$-th track. The model is trained to reconstruct a separate set of query trajectories $\mathbf{Q} = {q_{t,j}}$, which are randomly sampled from the video.

For each trajectory $j$, we embed its 3D positions $(x_{t,j}, y_{t,j}, z_{t,j})$ together with time $t$ using sinusoidal encoding (denoted by $\mathrm{PE}_{t,j}$). In parallel, we sample DINOv2 \cite{oquab2023dinov2} embeddings $f^{\text{DINO}}_{t,j}$ from the corresponding video frame regions. These two representations are 
concatenated as $e_{t,j} = [\mathrm{PE}_{t,j} \,\|\, f^{\text{DINO}}_{t,j}],$ and then projected into $C$ channels.  

A learnable ``readout" token is initialized randomly and appended, and self-attention is applied across all the tokens with an occlusion-aware mask $(1 - o_{t,j})$ to ignore hidden points. After attention, only the readout token is retained, producing a compact $C$-dimensional descriptor for the track. To integrate information across tracks, we adopt a Perceiver-style transformer \cite{jaegle2021perceiver} where a set of 128 latent tokens cross-attend to all track descriptors and then interact through self-attention. Finally, the latent tokens are compressed to yield a fixed $128 \times 64$ representation $\phi_S$, capturing both motion dynamics and semantic appearance cues. 

The decoder in \model operates on a motion latent $\phi_S$ that integrates 3D dynamics and semantic context. We train the decoder to reconstruct held-out query tracks. Concretely, given $\phi_S$ and a query point $(x_q, y_q, z_q, t_q)$, the decoder predicts the full trajectory passing through that point.  We first up-project the latent tokens in $\phi_S$ with an operator $U$ and add a query \emph{readout} token obtained from a sinusoidal encoding of $(x_q, y_q, z_q, t_q)$. Self-attention is applied over all tokens, after which only the readout token is retained. A final linear projection maps this token to the predicted trajectory $(\hat{x}_{t}^q, \hat{y}_{t}^q, \hat{z}_{t}^q, \hat{o}_{t}^q)$ across all frames.

\subsubsection{Training} 
 Following CoTracker3 \cite{karaev2024cotracker3}, we train \model on a combination of synthetic and real datasets to ensure both controlled supervision and real-world generalization. We use the Kubric3D dataset generator \cite{greff2022kubric} to create 
38k synthetic scenes with ground-truth 3D trajectories. While Kubric3D directly provides $(x_{t,j}, y_{t,j}, z_{t,j})$ for every point $j$ at time $t$, 
it does not include explicit occlusion labels. To obtain occlusion flags, we project each 3D point into the image plane and compare its depth  $z_{t,j}$ against the rendered depth map $D_t(x_{t,j}, y_{t,j})$ at that pixel. For depth maps we use VideoDepthAnything (VDA) metric model \cite{chen2025video}. The occlusion flag is then defined as $o_{t,j} = \mathbf{1}\!\left[\, z_{t,j} > D_t(x_{t,j}, y_{t,j}) + \epsilon \,\right]$, where $\epsilon$ is set to a small tolerance of 1e-4 to account for numerical precision. Thus, $o_{t,j} = 1$ indicates that the point is occluded at time $t$. In addition, we use the TAPVid-3D dataset \cite{koppula2024tapvid}, a large benchmark covering diverse real-world scenarios. TAPVid-3D contains  4,569 videos in the main split and 150 videos in the minival split, with  lengths ranging from 25 to 300 frames. Unlike Kubric3D, TAPVid-3D provides 
full ground-truth annotations, including 3D point trajectories as well as occlusion flags $o_{t,j}$, making it directly suitable for training and evaluating models under realistic settings. In our setup, we train on the main set and use the minival split for evaluation. 

During training, we randomly divide the point trajectories in each video into two equal parts. The first half is used as support tracks, which are passed through the encoder to produce the motion latents. The second half is used as query tracks, which the decoder must reconstruct given only the motion latents. 

We trained \model with the AdamW \cite{loshchilov2017decoupled} optimizer using a learning rate of 1e-4 warm-up followed by cosine decay. We initialize \model with the pretrained TRAJAN checkpoint and train for 300 epochs. Depth predictions are regularized with a scale-invariant penalty to handle differences in scale between synthetic and real domains. The DINO module \cite{oquab2023dinov2} is frozen during training. Additional training details, including hyperparameter settings, batch sizes, and ablation studies, are provided in Appendix~\ref{appendix:training}. The training loss is:
\begin{equation}
\mathcal{L}_{\text{model}}
= \sum_{q,t}\!\left(
w_{l1}\,\|\mathbf{p}_{q,t}-\hat{\mathbf{p}}_{q,t}\|_1
+ w_{\mathrm{BCE}}\,\mathrm{BCE}(o_{q,t},\hat{o}_{q,t})
\right).
\end{equation}

where $\mathbf{p}_{q,t} =(x_{q,t},y_{q,t},z_{q,t})$.
\begin{table*}[!t]
\centering
\resizebox{0.9\textwidth}{!}{%
\begin{tabular}{lcccccccccccc}
\toprule
\multirow{2}{*}{Method} & 
\multicolumn{3}{c}{\textbf{Aria}} & 
\multicolumn{3}{c}{\textbf{DriveTrack}} & 
\multicolumn{3}{c}{\textbf{PStudio}} & 
\multicolumn{3}{c}{\textbf{Average}} \\
\cmidrule(lr){2-4}\cmidrule(lr){5-7}\cmidrule(lr){8-10}\cmidrule(lr){11-13}
 & $AJ \uparrow$ & $APD \uparrow$ & $OA \uparrow$
 & $AJ \uparrow$ & $APD \uparrow$ & $OA \uparrow$
 & $AJ \uparrow$ & $APD \uparrow$ & $OA \uparrow$
 & $AJ \uparrow$ & $APD \uparrow$ & $OA \uparrow$ \\
 \midrule
TAPIR + ZD        & 15.7 & 23.5 & 79.8 & 6.3  & 10.5 & 81.6 & 11.2 & 18.9 & 78.7 & 11.0 & 17.6 & 80.1 \\
CoTracker + ZD    & 17.0 & 25.7 & 88.0 & 6.0  & 10.9 & 82.6 & 11.4 & 19.9 & 80.0 & 11.4 & 18.8 & 83.5 \\
BootsTAPIR + ZD   & 11.8 & 16.3 & 86.7 & 6.4  & 10.9 & 85.3 & 11.6 & 19.6 & 82.6 & 11.6 & 18.9 & 84.9 \\
CoTracker3 + ZD  & 15.6 & 24.1 & 88.6 & 13.3 & 19.6 & 86.8 & 9.0  & 13.6 & 83.9 & 12.6 & 19.1 & 86.4 \\
SpatialTracker & 16.7 & 25.7 & 89.3 & 6.9  & 12.4 & 83.7 & 12.3 & 21.6 & 78.5 & 12.0 & 19.9 & 83.8 \\
SpatialTrackerV2 & 18.6 & 26.3 & 90.8 & 16.4 & 24.3 & 90.2 & 18.1 & 27.6 & 86.7 & 17.7 & 26.0 & 89.2 \\
\midrule
CoTracker3-FT + ZD & 16.8 & 25.5 & 89.6 & 13.6 & 19.9 & 88.7 & 10.1  & 14.3 & 87.1 & 13.5 & 19.9 & 88.5 \\
\midrule
Ours & 17.7 & 24.9 & 89.2 & 11.9 & 14.8 & 85.7 & 12.3 & 19.9 & 82.5 & 14.0 & 19.8 & 85.8 \\
\bottomrule
\end{tabular}
}
\vspace{2mm}
\caption{\textbf{3D point tracking results on the TapVid-3D \emph{minival} set.} 
\model achieves competitive 3D tracking performance across datasets and performs on par with a finetuned CoTracker3 (CoTracker3-FT+ZD), highlighting its ability to reconstruct consistent and accurate 3D tracks despite the information bottleneck. \vspace{-1.5em}}
\label{tab:3dtracking}
\end{table*}

\subsubsection{Inference}
During inference, we operate directly on 2D input videos but we require 3D point tracks. Dense 2D point tracks $(x_{t,j}, y_{t,j})$ with occlusion are first estimated using CoTracker3 \cite{karaev2024cotracker3}, and subsequently lifted into 3D with metric depth predictions from VideoDepthAnything (VDA) \cite{chen2025video}. Half of the resulting 3D tracks $(x_{t,j}, y_{t,j}, z_{t,j})$ are then provided to the trained model as support tracks, with the other half being used as query tracks. The reconstructed query tracks produced by the decoder are finally used for evaluation.

For a lot of motion evaluation, we calculate the Average Jaccard ($AJ$) of the reconstructed tracks relative to the ``ground truth'' query tracks as a proxy for the reconstruction error. Following TAPVid-3D \cite{koppula2024tapvid}, as $AJ$ increases, the quality of reconstruction increases and vice-versa. The metric calculates the number of true positives (number of points whose spatial positions are within a $\delta_{3D}$ threshold, predicted correctly to be visible), divided by the sum of true positives and false positives (predicted
visible, but are occluded or farther than the threshold) and false negatives (visible points, predicted occluded or predicted to exceed the threshold).

\section{Results}\label{sec:results}
To demonstrate that \model can capture realistic, physical motion, we evaluate three complementary axes: its accuracy in 3D point tracking as described in Section \ref{sec:3d}, its ability to detect physical law violations in possible vs. impossible video pairs as described in Section \ref{sec:real_unreal}, and its alignment with human judgments of realism in generated videos as described in Section \ref{sec:ind}.

\vspace{-2mm}
\subsection{Can \model reconstruct 3D point tracks?}\label{sec:3d}
We evaluate \model on the TAPVid-3D minival set and report three 3D point tracking metrics: Occlusion Accuracy ($OA$), which measures the precision of occlusion predictions; $APD$ (in 3D), the average percentage of errors in 3D space within multiple threshold scales $\delta$; and Average Jaccard ($AJ$) (in 3D), which quantifies the accuracy of both position and occlusion estimation in 3D space. All metrics are taken from \citet{koppula2024tapvid}.

Since \model is an \emph{autoencoder} of point tracks, and therefore inherently less accurate due to its information bottleneck, we do not expect its performance in 3D point tracking to rival state-of-the-art approaches. Nevertheless, it is important that \model can reasonably accurately reconstruct 3D point tracks. We therefore compare \model against 3D-lifted versions of state-of-the-art 2D tracking methods lifted by Zerodepth (ZD) and 3D tracking methods like SpatialTracker \cite{xiao2024spatialtracker} and SpatialTrackerv2 \cite{xiao2025spatialtrackerv2}. Since most of these models were originally trained on the synthetic Kubric3D \cite{greff2022kubric} dataset, while our training data combines both Kubric3D (synthetic) and TAPVid-3D (real), we additionally fine-tune the CoTracker3 model \cite{karaev2024cotracker3} on TAPVid-3D pseudo labels (CoTracker-3 FT) and evaluate all models on the minival set. Table~\ref{tab:3dtracking} summarizes the comparative 3D tracking performance. \model consistently outperforms most baselines and achieves performance on par with CoTracker3 \cite{karaev2024cotracker3} when fine-tuned on the TAPVid-3D main dataset. 

We additionally provide an example of how \model performs in 3D track reconstruction when only a 2D input video is provided in Figure \ref{fig:3dreconstruct}. Despite the noisy depth signal obtained from VideoDepthAnything, \model reconstructs smooth 3D tracks for the generated video, and detects violations of physics when the hammer morphs inappropriately. 

Both results demonstrate that \model is capable of reconstructing 3D point tracks accurately despite its compressed latent space bottleneck, and motivates its candidacy as an automated metric for video realism.

\subsection{Can \model detect physical rule violations?}\label{sec:real_unreal}
For an automated metric of video realism to be useful, we need to be sure that it will detect physical rule violations.
To assess whether \model can reliably distinguish physically real and unreal scenarios, we evaluate on the IntPhys2 dataset \cite{bordes2025intphys}. IntPhys2 contains 1,012 videos across 253 scenes, organized as quadruplets of two \textbf{possible} (real) and two \textbf{impossible} (unreal) outcomes. Each video tests one of four core physical principles: \textbf{object permanence}, where objects continue to exist even when occluded; \textbf{object immutability}, where objects maintain their shape and structure; \textbf{spatio-temporal continuity}, where objects move smoothly through time and space; and \textbf{solidity}, where objects occupy space and cannot pass through one another. All videos are rendered in the Unreal Engine with both static and moving cameras.

\textbf{Baselines and ablations} We compare \model against several state-of-the-art vision-language models and self-supervised vision foundation models (baselines). We also compare \model to ablations which remove dimensional and semantic information. \textbf{\model (no DINO)} is a 3D extension of TRAJAN where we add an extra spatial dimension in the autoencoder to better capture motion dynamics. \textbf{\model (no 3D)} instead augments the 2D representation from TRAJAN with semantic features from DINOv2, while excluding 3D information and depth cues. \textbf{\model (no 3D, no DINO)} is identical to TRAJAN, and taken as the pre-trained checkpoint from \citet{allen2025direct}. Together, these variants highlight the individual roles of 3D structure and semantic context in detecting physical rule violations.

\begin{table*}[!t]
\centering
\resizebox{0.95\linewidth}{!}{%
\begin{tabular}{lcccccccc}
\toprule
\multirow{2}{*}{\textbf{Model / Category}} & 
\multicolumn{2}{c}{\textbf{Permanence}} & 
\multicolumn{2}{c}{\textbf{Immutability}} & 
\multicolumn{2}{c}{\textbf{Continuity}} & 
\multicolumn{2}{c}{\textbf{Solidity}} \\
\cmidrule(lr){2-3}\cmidrule(lr){4-5}\cmidrule(lr){6-7}\cmidrule(lr){8-9}
 & Fixed & Moving & Fixed & Moving & Fixed & Moving & Fixed & Moving \\
\midrule
GPT-4o \cite{hurst2024gpt} & 59.62 & 58.82 & 58.65 & 59.56 & 54.81 & 57.35 & 56.73 & 55.32 \\
Qwen-VL 2.5 \cite{bai2025qwen2} & 53.85 & 54.41 & 56.73 & 53.68 & 52.88 & 54.41 & 50.96 & 51.06 \\
Gemini-1.5 Pro \cite{gemini2024pro}  & 55.77 & 55.88 & 56.73 & 56.73 & 54.80 & 54.80 & 56.73 & 56.73 \\
Gemini-2.5 Flash \cite{gemini2025flash}  & 64.42 & 58.82 & 59.62 & 63.97 & 54.81 & 55.15 & 55.77 & 56.38 \\
VideoMAEv2-g \cite{wang2023videomae} & 63.46 & 50.00 & 54.81 & 53.69 & 65.38 & 54.41 & 48.08 & 59.57 \\
Cosmos-4B  \cite{agarwal2025cosmos}   & 51.92 & 41.18 & 50.96 & 48.32 & 53.85 & 50.00 & 48.08 & 55.32 \\
V-JEPA 2-h \cite{assran2025v} & 63.46 & 67.65 & 51.92 & 56.38 & 50.00 & 57.35 & 50.00 & 52.13 \\
V-JEPA-h+RoPE \cite{bardes2024revisiting}     & 59.62 & 57.35 & 55.77 & 58.72 & 57.69 & \textbf{75.00} & 46.15 & 58.51 \\
\midrule
\model (no 3D, no DINO) & 44.23 & 50.00 & 54.81 & 58.82 & 53.85 & 48.53 & 46.15 & 52.13 \\
\model (no 3D) & 61.54 & \textbf{76.47} & \textbf{75.00} & 73.08 & \textbf{78.85} & 73.53 &  69.23 & 59.57 \\
\model (no DINO) & 65.38 & 60.29 & 46.15 & 66.18 & 50.00 & 58.82 & 38.46 & 39.36 \\
\model & \textbf{76.92} & 75.00 & 73.08 & \textbf{76.47} & 67.31 & 69.12 & \textbf{70.77} & \textbf{64.47} \\
\midrule 
Human & 100.0 & 99.26 & 97.11 & 90.44 & 99.04 & 94.44 & 96.15 & 95.21 \\
\bottomrule
\end{tabular}
}
\vspace{2mm}
\caption{\textbf{Win rates (\%) on IntPhys2 across physical principles.} The top rows report prior models’ win rates, while the bottom rows compare \model, its ablations, and human performance. \model and \model (no 3D) achieve the strongest performance across most concept categories in detecting physically implausible events. \vspace{-1.5em}}
\label{tab:intphys2_combined}
\end{table*}

\vspace{0.5em}
\noindent \textbf{Performance analysis.}  
Table \ref{tab:intphys2_combined} shows the performance of \model against the baselines reported in \citet{bordes2025intphys} as well as our ablations. \model and \model (no 3D) significantly outperform all alternatives across most concept categories. Perhaps most surprisingly, \model shows the \emph{most benefit} over alternatives in the permanence ($\>+10\%$), immutability ($\>+10\%$), and solidity ($\>+5\%$) concept categories rather than continuity (approximately $-5 \; \text{to} \; +2\%$). This suggests that a small amount of 3D point track data is sufficient for models to learn what is physically plausible or not, and that reconstructing semantic 3D tracks is a better signal for learning \emph{realistic, plausible} physical motion than next frame prediction or next token prediction.

Most of the benefits of \model in determining possible vs. impossible physics may be due to the inclusion of DINOv2 features. Although \model performs best overall, the ablation which removes 3D structure but maintains DINOv2 features performs comparably to \model in most concept categories, indicating that \emph{semantic information} is key for understanding physical principles. By comparison, the other ablations perform comparably to previously evaluated predictive and Multimodal LLM (MLLM) approaches. We provide additional results in Appendix~\ref{appendix:intphys2}.

\begin{figure*}[!t]
    \centering
    \includegraphics[width=0.99\linewidth]{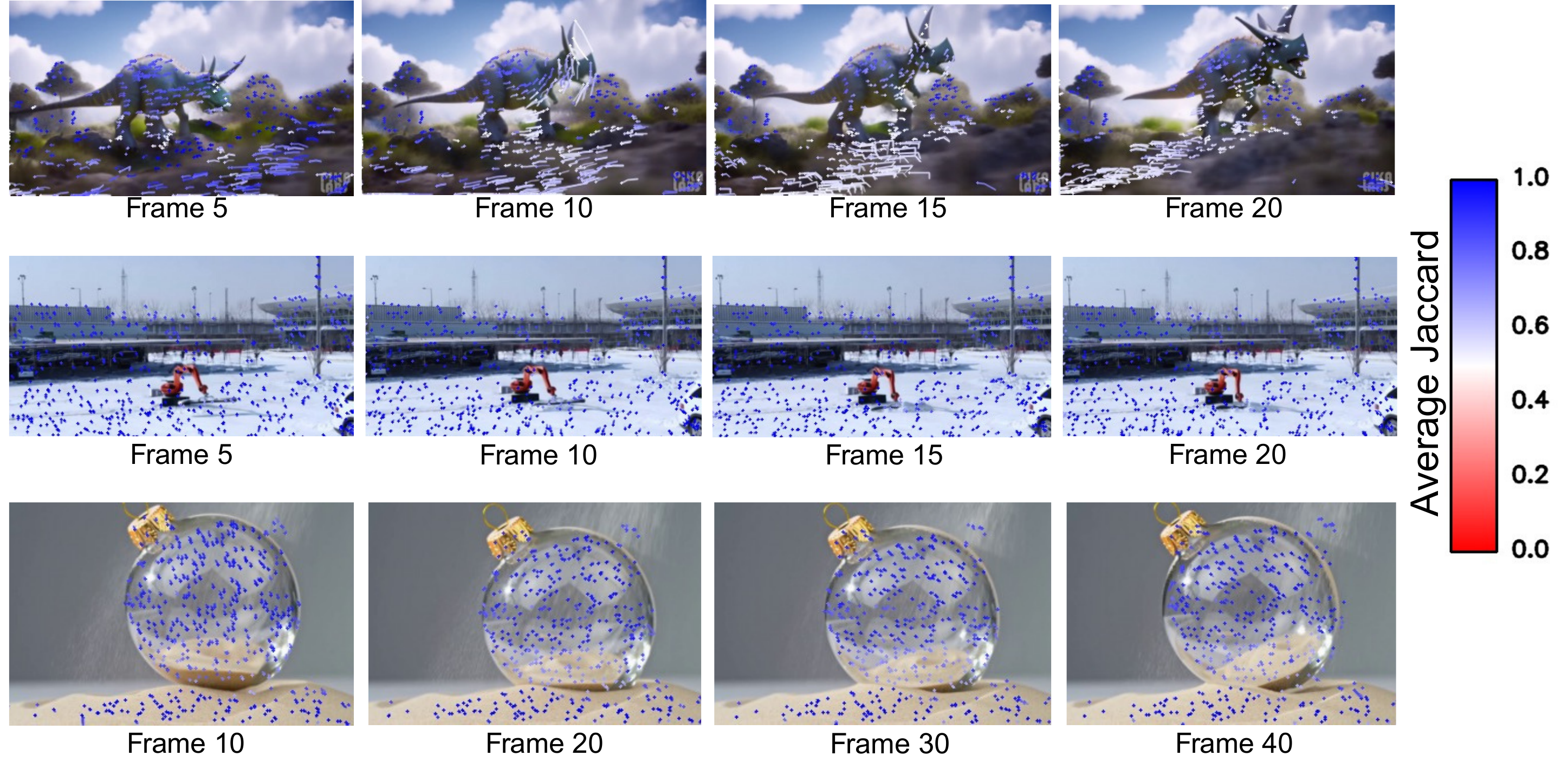}
    \caption{\textbf{Qualitative examples of accurate reconstruction on realistic generated videos from EvalCrafter and VideoPhy-2.} High-rated videos produce coherent and stable point tracks under \model (blue). The top and middle examples are from EvalCrafter~\cite{liu2024evalcrafter}, and the bottom example is from VideoPhy-2~\cite{bansal2025videophy}. \vspace{-1em}}

    \label{fig:3dspagood}
\end{figure*}

\subsection{Does \model capture human evaluations of realism in generated videos?}\label{sec:ind}
A key challenge in evaluating generated videos is measuring realism without relying on reference videos. This is particularly relevant when training data are inaccessible or when sampling a large number of outputs is computationally prohibitive. Human evaluation has therefore become the gold standard, since people can naturally judge whether motion appears realistic and physically plausible. 

\begin{figure*}[t]
    \centering
    \includegraphics[width=0.99\linewidth]{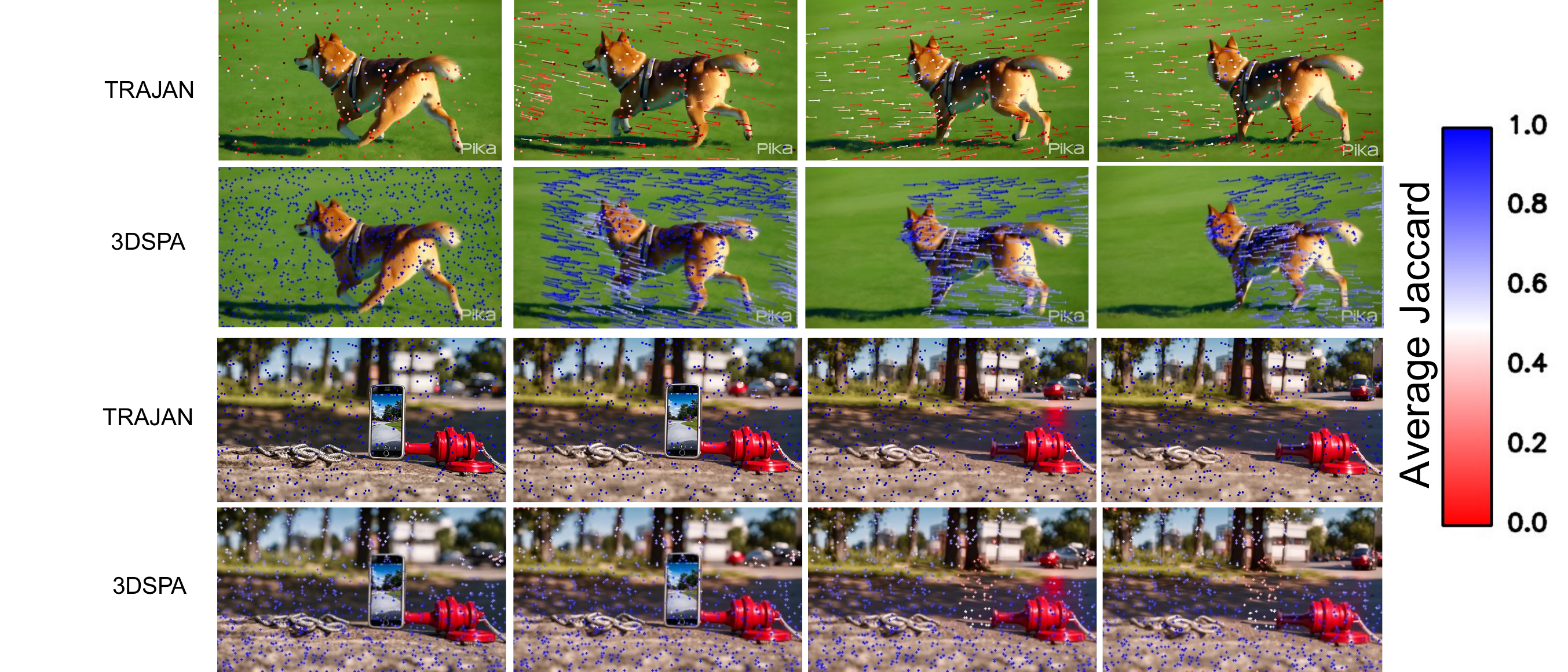}
    \caption{\textbf{TRAJAN vs. \model.} Qualitative comparison on videos from the EvalCrafter dataset \cite{liu2024evalcrafter}. Compared to TRAJAN \cite{allen2025direct}, \model produces more coherent and temporally stable point tracks, aligning more closely with human judgments of motion quality. In the top example (dog walking; human rating: 4.5/5), \model accurately captures articulated leg motion in 3D, whereas TRAJAN produces noisy and inconsistent tracks. In the bottom example (human rating: 1.67/5), the phone gradually disappears; \model correctly identifies this semantic violation, while TRAJAN fails due to smooth but semantically implausible trajectories.}
    \label{fig:trajan3dspa}
\end{figure*}

We draw on two datasets which include a large set of videos generated by multiple generative video models: VideoPhy-2 \cite{bansal2025videophy} and EvalCrafter \cite{liu2024evalcrafter}.
VideoPhy-2 emphasizes action-centric videos and includes human annotations of physical commonsense and semantic adherence to the text prompt. EvalCrafter \cite{liu2024evalcrafter} evaluates video quality with a larger set of five metrics including motion quality, temporal consistency, and several prompt adherence measures. \autoref{fig:3dspagood} shows representative high-rated videos from both datasets, where \model produces coherent and temporally stable point tracks, indicating strong alignment with human judgments of motion realism and physical plausibility. These videos were all rated highly for realistic motion and cover challenging situations for point tracking including transparency and substantial motion. In all cases, \model reconstructs the point tracks well, since the videos appear realistic.  We show more examples in Appendix \ref{appendix:badmotion}.

\textbf{VideoPhy-2 } We use the VideoPhy-2 benchmark \cite{bansal2025videophy} to assess how well \model performs as an automated realism metric relative to human judgments. This benchmark emphasizes two key aspects: \textit{semantic adherence} (SA), which measures whether generated videos follow the intended action semantics, and \textit{physical commonsense} (PC), which evaluates whether the motion and interactions in videos are consistent with intuitive physical rules. We are primarily interested in physical commonsense. \citet{bansal2025videophy} also provide an automated evaluation metric, VIDEOPHY-2 AutoEval, which is a vision-language model fine-tuned to predict a physical commonsense score on a subset of the generated video dataset.

We measure the automated metric quality by correlating model ratings and human ratings with the Spearman rank coefficient. Model ratings are calculated automatically as the  Average Jaccard for each video -- a proxy for the reconstruction error of the autoencoder.

As shown in Table \ref{tab:videophy2_spearman}, \model substantially outperforms 2D variants in tracking human ratings of physical commonsense and also provides a significant boost over the 3D baseline. The inclusion of both 3D structural cues and semantic DINO features enables stronger alignment with human assessments, where \model achieves the highest Spearman rank coefficient among ablations.
More remarkably, \model strongly outperforms most vision-language models (VideoCon, VideoScore, and VideoLlava) on this task, and even closely matches VIDEOPHY-2 AutoEval despite not being trained on the provided dataset. We additionally analyze the computational cost of \model and representative baselines. \model achieves comparable or faster inference time relative to VLM-based evaluators, with approximately a 2x overhead relative to TRAJAN. Detailed runtime breakdowns are provided in \autoref{app:inference_time}.

\begin{table}[!tbph]
    \centering
    \begin{tabular}{lc}
\toprule
\textbf{Model} & \textbf{Spearman (PC)} \\
\midrule
\multicolumn{2}{c}{\textit{Video Evaluation Models (fine-tuned)}} \\
VideoCon–Physics    & 0.48 \\
VideoCon            & 0.13 \\
VideoLlava          & 0.08 \\
VideoScore          & 0.17 \\
VIDEOPHY–2–AUTOEVAL & \textbf{0.76} \\
\midrule
\multicolumn{2}{c}{\textit{Ablations (no fine-tuning)}} \\
\model (no 3D, no DINO) & 0.19 \\
\model (no 3D)         & 0.40 \\
\model (no DINO)       & 0.50 \\
\model (ours)          & \textbf{0.74} \\
\bottomrule
\end{tabular}
\caption{Spearman rank coefficients on the VideoPhy-2 benchmark for physical commonsense (PC). Video Evaluation Models are fine-tuned vision-language models. \vspace{-1em}}
\label{tab:videophy2_spearman}
\end{table}

\begin{table*}[h]
\centering
\begin{tabular}{lccccc}
\toprule
\textbf{Model} & \textbf{Visual Quality} & \textbf{T2V} & \textbf{Motion Quality} & \textbf{Consistency} & \textbf{Subjective Likeness} \\
\midrule
\model (no 3D, no DINO)  & 0.28 & 0.25 & 0.24 & 0.33 & 0.23 \\
\model (no 3D) & 0.31 & 0.44 & 0.41 & 0.39 & 0.46 \\
\model (no DINO) & 0.45 & 0.55 & 0.49 & 0.48 & \textbf{0.63} \\
\model (ours) & \textbf{0.48} & \textbf{0.58} & \textbf{0.55} & \textbf{0.60} & 0.60 \\
\bottomrule
\end{tabular}%
\caption{Spearman rank coefficients between human annotations and automated AJ scores for the EvalCrafter \cite{liu2024evalcrafter} dataset in the categories of visual quality, text-to-video similarity, motion quality, temporal consistency, and subjective likeness across different ablations. \vspace{-2em}}
\label{tab:evalcrafter}
\end{table*}

\textbf{EvalCrafter } Similarly to VideoPhy-2, EvalCrafter \cite{liu2024evalcrafter} is a dataset consisting of generated videos from several frontier generative video models. For each video, a set of human annotators rated the \textbf{visual quality}, \textbf{text to video consistency}, \textbf{motion quality}, \textbf{temporal consistency}, and \textbf{subjective likeness}. Similarly to VideoPhy-2, we compute Spearman rank coefficients between human ratings and the Average Jaccard (AJ) for each of the ablations and \model. Since many videos in EvalCrafter contain no motion (and therefore could not be assessed as physically realistic or not), we restricted evaluation to videos with medium to high motion, defined as the top 50\% of videos ranked by change in 3D point track positions, yielding a test set of 1,849 videos. Table~\ref{tab:evalcrafter} clearly demonstrates that \model achieves the best performance; further highlighting that integrating both 3D structure and semantic DINO features provides the strongest predictor of a variety of human annotations for generated videos. 

Figure \ref{fig:trajan3dspa} shows two example videos which highlight the differences between \model and TRAJAN (i.e. the \model (no 3D, no DINO) ablation) in capturing motion quality, the closest metric to physical realism. In the first video, a dog is walking on grass. Here, the 3D nature of the dog's legs is critical for modeling its motion. Without 3D structure, TRAJAN is unable to correctly capture this motion, and incorrectly labels the video as unrealistic. In the second video, a phone slowly disappears. Since TRAJAN has no notion of object semantics, and the point trajectories are smooth (due to the \emph{slow} disappearance), it incorrectly believes this video is realistic. \model instead understands that phones cannot simply disappear, and marks this as an unrealistic video. In both cases, this matches human raters' intuitions.

\section{Discussion}
We introduce the 3D Semantic Point Autoencoder (3DSPA), a model for evaluating video realism using semantic-aware 3D point trajectories. Across several experiments, we found that \model's combination of semantic and 3D geometric information was crucial for (1) 3D point track reconstruction, (2) physical rule violation detection, and (3) matching various human annotations of generated videos, including motion quality and adherence to physical commonsense. Through ablation studies, we determined that semantic information is particularly crucial to determining whether a video is physically realistic -- geometry alone is not enough. Perhaps most surprisingly, \model outperforms state-of-the-art vision-language models both in detecting synthetic physical rule violations such as solidity and immutability, and in tracking human physical commonsense judgments of generated videos. 

However, \model also has some limitations. 
For example, \model's 3D point track reconstruction can be unreliable (see Appendix~\ref{appendix:reconstructfails}), especially in complex scenes where depth cues are difficult to interpret and VideoDepthAnything \cite{chen2025video} performs poorly. This in turn can propagate errors into downstream realism scores. While our results indicate that \model remains robust to these artifacts and performs well on average across diverse environments, these limitations nevertheless motivate future work on further improving the robustness of trajectory reconstruction. 

Overall, \model offers a scalable alternative to human evaluation of video realism. We believe that 3D point tracks naturally capture depth-aware motion, interactions, and occlusion, making them more effective than frame-based metrics for spotting subtle physics violations. In future work, we plan to make trajectories depend on past motion, enabling stronger tests of long-term dynamics and temporal realism, as well as investigating whether these metrics can be used to improve or regularize the training of generative video models.

\section{Acknowledgements}
KA is supported by a Canada CIFAR AI Chair. Resources used in preparing this research were provided, in part, by the Province of Ontario, the Government of Canada through CIFAR, and companies sponsoring the Vector Institute.

\section{Broader Impacts}
This paper presents a method to help identify errors in generated videos. It may be used to assist in the detection of fake content, but may also be extended to make generated videos more realistic. This could have both positive and negative societal consequences --- better detection of fake content, and more convincing fake content. There are more potential societal consequences of our work, but these do not need to be specifically highlighted here.

\bibliography{example_paper}
\bibliographystyle{icml2026}

\newpage
\appendix
\onecolumn

\section{Training Setup \& Hyperparameters} \label{appendix:training}
We train our model using AdamW \cite{loshchilov2017decoupled} with a cosine learning rate schedule, preceded by a warmup of 10000 steps. The peak learning rate is set to $1 \times 10^{-4}$. Training is performed for 300 epochs with a batch size of 256. This extended training schedule, along with the larger batch size, allows the model to better stabilize its motion representation and improve generalization across diverse video scenarios. 

As before, we supervise both position and occlusion prediction, applying a L1 loss on $(x_t, y_t)$ coordinates and a cross-entropy loss on the occlusion logit $o_t$. We maintain the weighting ratio of $5000:10^{-8}$, which prioritizes motion fidelity while encouraging invariance to occlusion. We found that balancing these losses equally degraded correlation with human judgments of realism, consistent with our earlier observations.

To improve temporal localization of query points, we replace the naive linear up-projection operator with a strided-window upsampling operator. Specifically, each latent token $\phi^l_S$ is linearly up-projected and concatenated with a temporal window $[\rho_t : \rho_t+128)$ along the channel axis. This encourages the decoder to attend to temporally relevant information for a given query point. 

\subsection{Hyperparameters}
Tables~\ref{tab:trajan_proj_hparams} and \ref{tab:trajan_transformer_hparams} provide the full set of hyperparameters for positional encoding, projection operators, and transformer modules. Compared to the original TRAJAN configuration, we increase the dimensionalities of the projection layers as we have increased the data and added additional parameters due to DINO and depth features, enabling richer multi-modal fusion of semantic and geometric cues.

\begin{table}[h]
\centering
\resizebox{0.8\textwidth}{!}{%
\begin{tabular}{lc}
\toprule
\textbf{Component} & \textbf{Hyperparameter Value} \\
\midrule
Sinusoidal embedding (spatial + temporal + depth) & 32 frequencies \\
Track token projection dimensionality ($C$) & 384 \\
DINO feature projection dimensionality & 768 \\
Depth feature projection dimensionality & 256 \\
Compression dimensionality & 96 \\
Up-projection dimensionality & 1280 \\
Query point encoder dimensionality & 1280 \\
\bottomrule
\end{tabular}}
\vspace{2mm}
\caption{Positional encoding and projection operator hyperparameters for \model.}
\label{tab:trajan_proj_hparams}
\end{table}

\begin{table}[t]
\centering
\resizebox{1.0\textwidth}{!}{%
\begin{tabular}{lccccc}
\toprule
\textbf{Transformer Name} & \textbf{Attention Type} & \textbf{QKV Size} & \textbf{Layers} & \textbf{Heads} & \textbf{MLP Size} \\
\midrule
Input 3D track transformer & SA & $96 \times 8$ & 3 & 8 & 1536 \\
Perceiver-style transformer & CA & $96 \times 8$ & 4 & 8 & 2048 \\
Up-projection latent transformer (decoder) & CA & $96 \times 8$ & 4 & 8 & 2048 \\
Track readout transformer & CA & $96 \times 8$ & 4 & 8 & 1536 \\
\bottomrule
\end{tabular}}
\vspace{2mm}

\caption{Transformer architecture hyperparameters for \model . SA = self-attention, CA = cross-attention.}
\label{tab:trajan_transformer_hparams}
\end{table}
\section{Additional results on Intphys2} \label{appendix:intphys2}
\noindent \textbf{Dataset Structure.} The videos are further categorized by difficulty:  
\begin{itemize}
    \item \textbf{Easy (104 videos):} Simple environments with colorful geometric shapes.  
    \item \textbf{Medium (400 videos):} Diverse backgrounds with textured shapes.  
    \item \textbf{Hard (336 videos):} Realistic objects within cluttered, complex backgrounds.
    \item \textbf{Unknown (172 videos):} Mixed or ambiguous scenes.
\end{itemize}

\begin{figure}[h]
    \centering
    \includegraphics[width=1.0\textwidth]{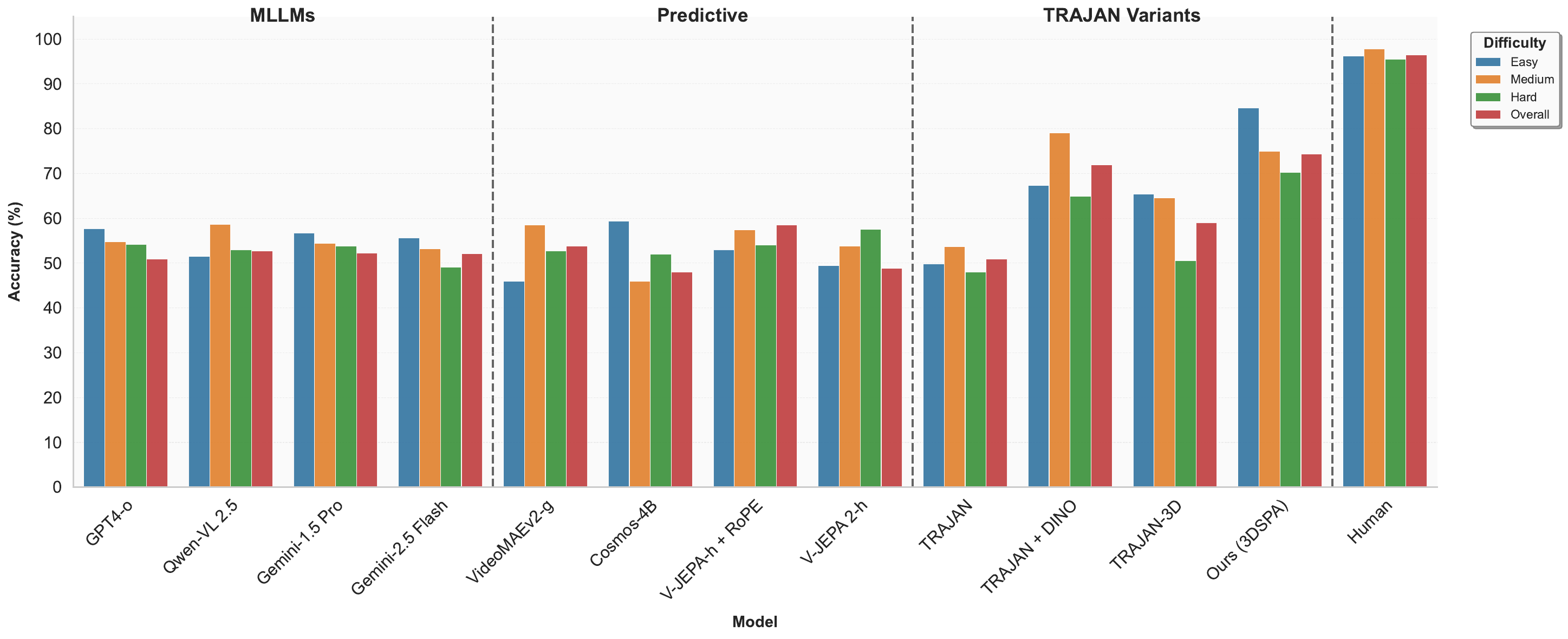}
    \caption{Performance comparison across models on the IntPhys2 benchmark for each of the \emph{easy}, \emph{medium} and \emph{hard} categories.} \label{fig:intphys_barplot}
\end{figure}

\noindent We report results across the three difficulty variants to better capture how model performance scales with increasing visual and physical complexity. This breakdown provides motivation for our evaluation, as it disentangles robustness to simple synthetic settings from generalization to realistic and cluttered environments (see Figure~\ref{fig:intphys_barplot}). Notably, performance on the \textbf{hard} category is consistently the lowest in terms of Average Jaccard, highlighting the challenge of reconstructing tracks in realistic scenes with heavy clutter, occlusions, and object interactions.

\begin{figure}
    \centering
    \includegraphics[width=0.99\linewidth]{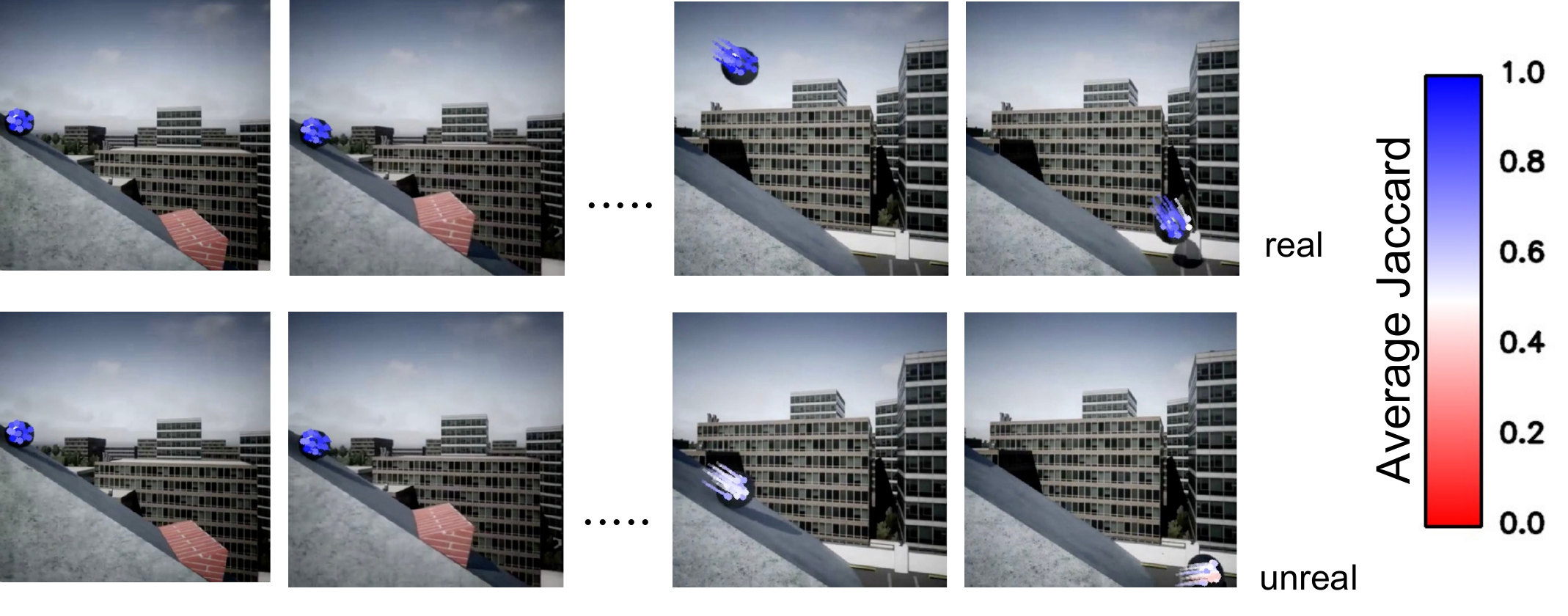}
\caption{\textbf{Visualization of point tracks on IntPhys2 videos.} For each video, we project tracked points and color them using their Average Jaccard (AJ) consistency score (red = low / unreal, blue = high / real). Real, physically consistent videos (top) show smooth, coherent tracks with high AJ scores throughout the video, while unreal videos with physical rule violations (bottom) exhibit poorly reconstructed point trajectories.}
    \label{fig:intphys2}
\end{figure}

\noindent We also give a qualitative example of how \model captures physical rule violations in \autoref{fig:intphys2}. Recall that the IntPhys2 dataset is constructed by pairing the same initial frames with either a realistic ending (real) or one which violates a physical rule (unreal). In \autoref{fig:intphys2}, the ball should interact with the ramp and fly through the air (top, real). In this case, the point tracks on the ball are well reconstructed. However, in the bottom video, where the ball just continues down the slope without flying through the air, the point tracks are poorly reconstructed.

\begin{table}[h]
\centering
\begin{tabular}{l c}
\toprule
\textbf{Model} & \textbf{Inference Time (s)} \\
\midrule
VideoCon--Physics & 8.13 \\
VideoCon & 6.21 \\
VideoLLaVA & 13.52 \\
VideoScore & 4.89 \\
VIDEOPHY--2--AUTOEVAL & 10.67 \\
\midrule
TRAJAN & 2.18 \\
\textbf{\model (ours)}  & \textbf{5.61} \\
\bottomrule
\end{tabular}
\vspace{2mm}

\caption{Average inference time per video for \model and TRAJAN as compared to other finetuned VLM models}
\label{tab:inference_time}
\end{table}

\section{Inference Time Comparison}
\label{app:inference_time}

We benchmark the inference cost of \model and representative baselines on a single NVIDIA A100-40GB GPU. All methods are evaluated on videos with 50 frames per clip, averaged over 50 different clips. Reported timings include full end-to-end processing required to produce a final evaluation score for a video. To Note: for \model, the pipeline includes four stages: (1) CoTracker3 for dense 2D point tracking, (2) Video Depth Anything (VDA) for per-frame depth estimation, (3) DINOv2 for semantic feature extraction, and (4) \model autoencoder inference for trajectory encoding and reconstruction. All components are executed sequentially on GPU without batch parallelization across videos.

Table~\ref{tab:inference_time} reports the average inference time per video. Despite incorporating multiple perception modules, \model takes almost similar time as most VLM-based evaluators while offering stronger sensitivity to physical realism and motion artifacts. Compared to TRAJAN, \model incurs approximately a 2$\times$ overhead due to depth and semantic feature extraction, which enables richer 3D semantic reasoning.

\section{\model successes and failures on 3D point track reconstruction} \label{appendix:reconstructfails}

\begin{figure}
    \centering
    \includegraphics[width=0.99\linewidth]{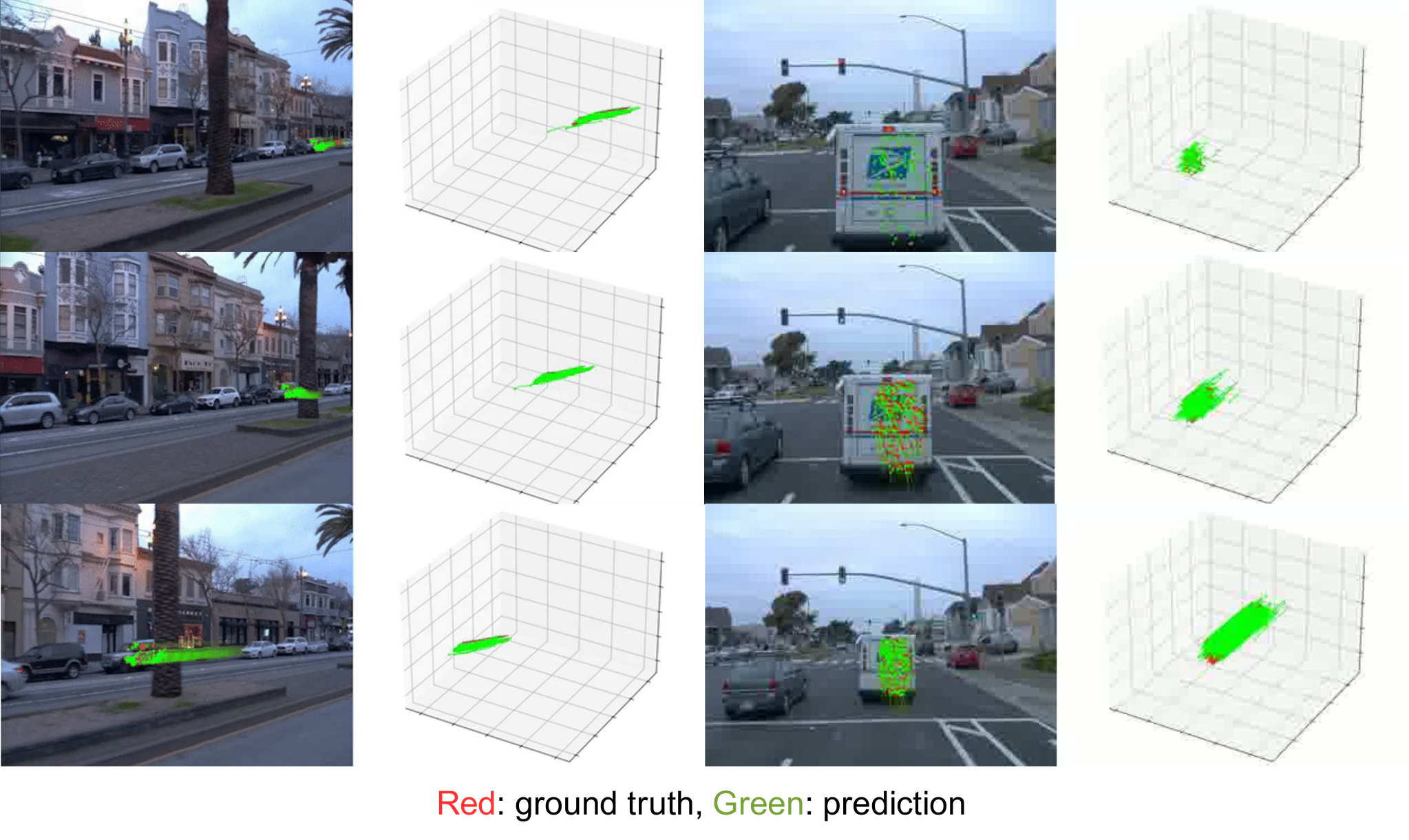}
    \caption{\textbf{Positive examples where \model performs well on TAP-Vid3D-Minival.}
Predicted 3D trajectories (green) align closely with ground truth (red) even for videos with small objects and high motion. These videos are from the DriveTrack dataset within TAP-Vid3D-Minival.}
    \label{fig:tapvid3dgood}
\end{figure}

\begin{figure}
    \centering
    \includegraphics[width=0.99\linewidth]{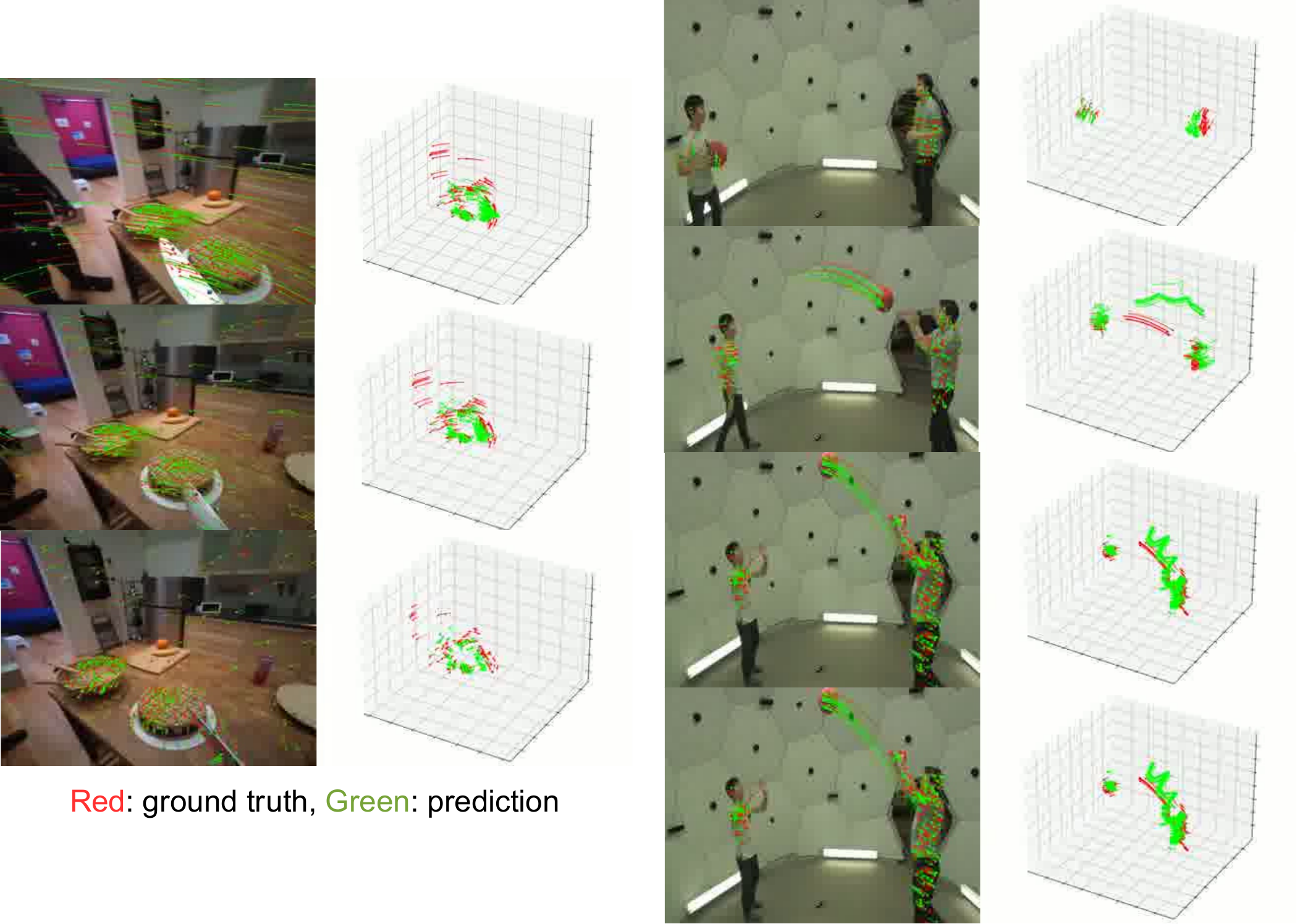}
    \caption{\textbf{Failure cases of \model on TAP-Vid3D-Minival.} We visualize ground-truth (red) and predicted (green) query 3D point trajectories for challenging scenes. Errors typically arise in regions with complex geometry or depth ambiguities, giving high error in depth prediction, although the motion following is good. Corresponding 2D frames (left) and reconstructed 3D trajectories (right) highlight mismatches between predicted and true motion, which mainly arise from incorrectly estimated depth.}
    \label{fig:tapvid3dbad}
\end{figure}

\noindent In this section, we give some examples of how well \model can reconstruct 3D point tracks on the TAP-Vid3D-Minival dataset. In \autoref{fig:tapvid3dgood}, we show successes of \model in reconstructing point tracks of both small objects and fast moving objects. However, \model struggles when depth cues are difficult to interpret (\autoref{fig:tapvid3dbad}). Since we rely on VideoDepthAnything~\cite{chen2025video}, when depth cues are ambiguous, we see poorer 3D tracking. However, the 2D tracks still look reasonable.

\section{Visualizations showing \model's ability to capture bad motion in generated videos}\label{appendix:badmotion}

\begin{figure}
    \centering
    \includegraphics[width=0.99\linewidth]{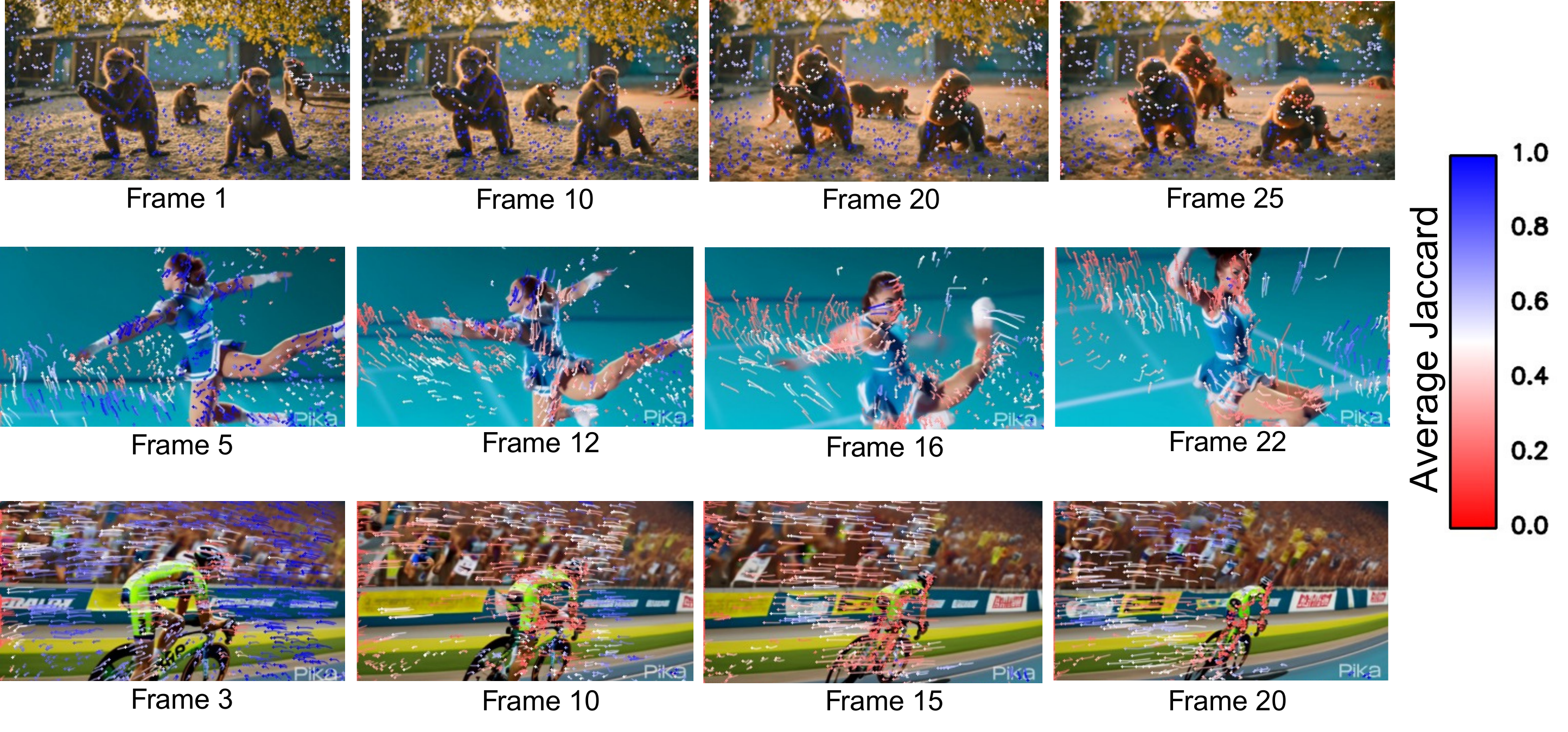}
    \caption{\textbf{Additional Visualization on Evalcrafter \& Videophy-2.} Low-rated motion videos lead to lower AJ scores and unstable point tracks from \model, reflecting difficulty in reconstructing unreliable motion. The top videos are from Evalcrafter~\cite{liu2024evalcrafter}, middle and bottom are from Videophy2~\cite{bansal2025videophy}.}
    \label{fig:3dspabad}
\end{figure}

\noindent Here we show \model's ability to capture unrealistic motion across a selection of generated videos. In \autoref{fig:3dspabad}, we instead show examples of videos that were rated \emph{poorly} for motion realism. In these cases, objects are morphing shape or inappropriately disappearing. The tracks are poorly reconstructed exactly where the motion becomes unrealistic.

\end{document}